\documentclass[letterpaper, 10 pt, journal, twoside]{format/IEEEtran}

\IEEEoverridecommandlockouts    
\usepackage{glossaries}
\newacronym{mav}{MAV}{Micro Aerial Vehicles}
\newacronym{uav}{UAV}{Unmanned Aerial Vehicle}
\newacronym{ovc}{OVC}{Open Vision Computer}
\newacronym{lidar}{LiDAR}{Light Detection and Ranging}
\newacronym{vio}{VIO}{visual-inertial odometry}
\newacronym{gpgpu}{GPGPU}{General-Purpose Graphics Processing Unit}
\newacronym{ugv}{UGV}{Unmanned Ground Vehicle}
\newacronym{uwb}{UWB}{Ultra Wideband}
\newacronym{svm}{SVM}{Support Vector Machine}
\newacronym{fcn}{FCN}{Fully Convolutional Network}
\newacronym{cnn}{CNN}{Convolutional Neural Network}
\newacronym{loam}{LOAM}{LiDAR Odometry and Mapping}
\newacronym{sloam}{SLOAM}{Semantic LiDAR Odometry and Mapping}
\newacronym{slam}{SLAM}{Simultaneous Localization and Mapping}
\newacronym{iot4ag}{IoT4Ag}{NSF Engineering Research Center for the Internet of Things for Precision Agriculture}
\newacronym{grasp-lab}{GRASP Lab}{the General Robotics, Automation, Sensing and Perception Laboratory}
\newacronym{jps}{JPS}{Jump Point Search}
\newacronym{ukf}{UKF}{Unscented Kalman Filter}
\newacronym{sam}{SAM}{Smoothing and Mapping}
\newacronym{icp}{ICP}{Iterative Closest Point}
\newacronym{imu}{IMU}{Inertial Measurement Unit}
\newacronym{tsdf}{TSDF}{Truncated Signed Distance Field}
\newacronym{esdf}{ESDF}{Euclidean Signed Distance Field}
\newacronym{sdf}{SDF}{Signed Distance Field}
\newacronym{rrt}{RRT}{Rapidly Exploring Random Tree}
\newacronym{fpv}{FPV}{First-person View}
\newacronym{dnn}{DNN}{Deep Neural Network}
\newacronym{igpred}{IGPred}{Information Gain Prediction}
\newacronym{csqmi}{CSQMI}{Cauchy-Schwarz Quadratic Mutual Information}
\newacronym{nbv}{NBV}{Next Best View}
\newacronym{vae}{VAE}{Variational Autoencoder}
\newacronym{tsp}{TSP}{Traveling Salesman Problem}
\newacronym{bcsm}{BCSM}{Behavior Control State Machine}
\newacronym{pca}{PCA}{Principal Component Analysis}
\newacronym{aspp}{ASPP}{Atrous Spatial Pyramid Pooling}
\newacronym{swap}{SWaP}{Size Weight and Power}
\newacronym{soi}{SoI}{Semantic Object of Interest}
\newacronym{aoi}{AoI}{Area of Interest}
\newacronym{drl}{DRL}{Deep Reinforcement Learning}
\newacronym{dl}{DL}{Deep Learning}
\newacronym{fov}{FoV}{Field of View}
\newacronym{tops}{TOPS}{Tera Operations per Second}

\usepackage[table,usenames,dvipsnames]{xcolor}      

\usepackage{tabularx}
\usepackage{multirow, multicol}
\usepackage{array}
\usepackage{tabu}

\usepackage{tikz}
\usepackage{pgfplots}
\usetikzlibrary{shapes.geometric, arrows}
\usepackage{float}

\newcolumntype{P}[1]{>{\centering\arraybackslash}p{#1}}
\newcolumntype{M}[1]{>{\centering\arraybackslash}m{#1}}
\usepackage{booktabs}
\newcolumntype{N}{>{\centering\arraybackslash}m{.5in}}
\newcolumntype{G}{>{\centering\arraybackslash}m{2in}}

\let\labelindent\relax

\usepackage{extarrows}                              
\usepackage{enumitem}
\usepackage{wrapfig}

\usepackage{amsmath,amssymb,amsfonts,amsthm,dsfont, bm} 
\usepackage{mathtools}
\usepackage{algorithm,algorithmicx,listings}        
\usepackage[noend]{algpseudocode}			              
\makeatletter
\def\BState{\State\hskip-\ALG@thistlm}
\makeatother
\DeclarePairedDelimiter\abs{\lvert}{\rvert}%
\DeclarePairedDelimiter\norm{\lVert}{\rVert}%
\makeatletter
\let\oldabs\abs
\def\abs{\@ifstar{\oldabs}{\oldabs*}}
\let\oldnorm\norm
\def\norm{\@ifstar{\oldnorm}{\oldnorm*}}
\makeatother

\usepackage{graphicx}
\usepackage{subcaption}
\usepackage{textcomp}
\usepackage[font=footnotesize]{caption}
\usepackage{subcaption}

\usepackage{xspace}
\usepackage{placeins}

\makeatletter
\DeclareRobustCommand\onedot{\futurelet\@let@token\@onedot}
\def\@onedot{\ifx\@let@token.\else.\null\fi\xspace}

\makeatother

\usepackage[normalem]{ulem}
\usepackage[utf8]{inputenc}
\usepackage[english]{babel}

\DeclareMathAlphabet\mathbfcal{OMS}{cmsy}{b}{n}

\newtheorem*{assumption*}{Assumption}

\newtheorem*{problem*}{Problem}

\usepackage{graphicx}

\usepackage{lipsum}
\makeatletter
\let\NAT@parse\undefined
\makeatother
\usepackage[space, compress, sort, noadjust]{cite}
\usepackage[breaklinks=true, colorlinks, bookmarks=true, citecolor=Black, urlcolor=Violet,linkcolor=Black]{hyperref}
\usepackage{balance}

\begin{document}

\title{
Gaussian Splatting as a Unified Representation for \\ 
Autonomy in Unstructured Environments

}

\author{Dexter Ong, Yuezhan Tao, Varun Murali, Igor Spasojevic,  Vijay Kumar, Pratik Chaudhari %
\thanks{
This work was supported by TILOS under NSF Grant CCR-2112665, IoT4Ag ERC under NSF Grant EEC-1941529, the ARL DCIST CRA W911NF-17-2-0181, DSO National Laboratories and NVIDIA.
All authors are with GRASP Laboratory, University of Pennsylvania
{\tt\footnotesize\{odexter, yztao, mvarun, igorspas, kumar, pratikac\}@seas.upenn.edu}.}
}

\maketitle

\begin{abstract}
In this work, we argue that Gaussian splatting is a suitable unified representation for autonomous robot navigation in large-scale unstructured outdoor environments.
Such environments require representations that can capture complex structures while remaining computationally tractable for real-time navigation.
We demonstrate that the dense geometric and photometric information provided by a Gaussian splatting representation is useful for navigation in unstructured environments.
Additionally, semantic information can be embedded in the Gaussian map to enable large-scale task-driven navigation.
From the lessons learned through our experiments, we highlight several challenges and opportunities arising from the use of such a representation for robot autonomy.

\end{abstract}

\section{Introduction}
\label{sec:intro}

\begin{figure}[!t]
\begin{subfigure}{0.2\textwidth}
    \includegraphics[height=3.7cm]{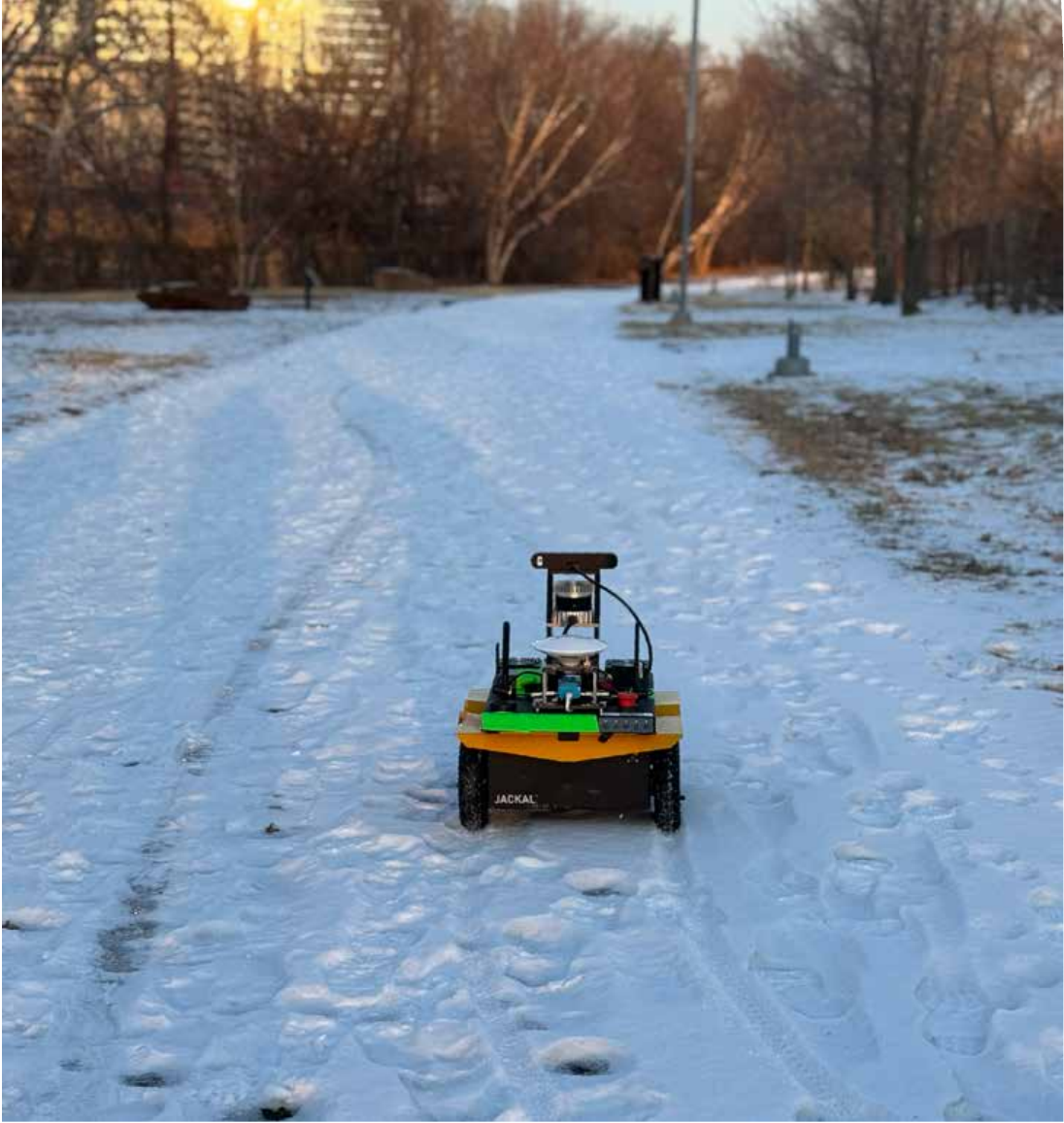}
\end{subfigure}
\hspace{-0.2cm}
\begin{subfigure}{0.2\textwidth}
    \includegraphics[height=3.7cm]{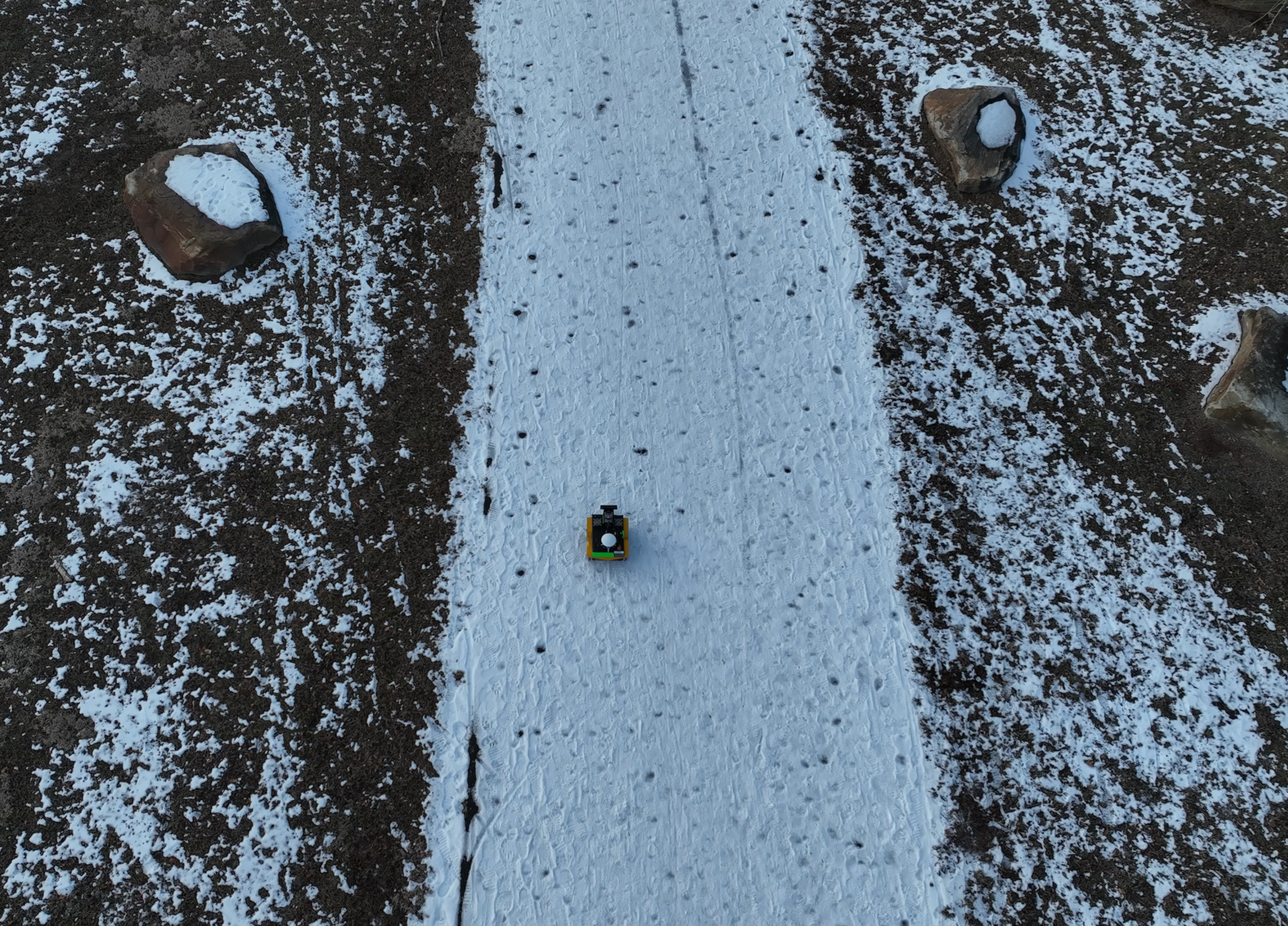}
\end{subfigure}
\caption{Robot navigating outdoor environments that lack structure in semantics and geometry.}
\label{fig:robot_snow}
\vspace{-0.5cm}
\end{figure}

Robots navigating large outdoor spaces with minimal visual structure face significant challenges in perception, mapping, and decision-making.
In environments such as those in Fig.~\ref{fig:robot_snow}, traditional approaches often struggle to capture the complexity and variability of the scene, presenting challenges for autonomous navigation under such conditions.
These capabilities are crucial for applications such as precision agriculture~\cite{9705188}, forestry~\cite{prabhu2024uavs}, search-and-rescue~\cite{tian2020search} and infrastructure inspection~\cite{bircher2018receding}.
To address this, we present Gaussian splatting as a versatile representation for large-scale autonomy in unstructured outdoor environments.

Gaussian splatting offers several advantages for representing unstructured environments.
Its dense representation can effectively capture thin objects and complex geometries and offers a good balance between dense representations like point clouds and more discretized structures found in mesh and voxel-based approaches. 
Moreover, Gaussian splatting supports representing the environment as a semantic feature field, enabling dense semantic features that facilitate clustering based on semantics rather than purely geometric information.
This capability is particularly valuable for real-time scene understanding in unstructured environments and for providing context for executing under-specified tasks in the wild where the maps are unknown \textit{a priori}.

In this paper, we demonstrate that Gaussian splatting can capture the necessary semantic and geometric information to serve as a unified representation for exploration, semantic navigation, and collision avoidance. We highlight four key capabilities of our Gaussian splatting-based approach:

\begin{enumerate}
    \item We introduce an efficient metric for estimating uncertainty in the Gaussian splatting map, enabling information-driven exploration of the environment. This allows robots to autonomously identify and investigate areas of high uncertainty, improving the completeness and accuracy of the map.
    \item We present a hierarchical mapping and planning framework built on language-embedded Gaussian splatting. This enables task-driven semantic navigation, allowing robots to interpret high-level commands and navigate based on semantic understanding of the environment.
    \item Our system leverages GPU-based collision checking in a dense Gaussian splatting map for trajectory planning, enabling efficient collision avoidance in complex environments.
    \item To address the challenges of mapping large-scale environments, we present a submapping framework for Gaussian splatting with particular considerations for autonomy. This framework enables efficient mapping and collision checking of Gaussians in expansive outdoor spaces, making our approach scalable to real-world applications.
\end{enumerate}
By integrating these capabilities, our Gaussian splatting-based approach provides a unified map representation for mapping, planning, and collision avoidance in large-scale and unstructured outdoor environments. 

\section{Related Work}
\label{sec:related work}

\subsection{Map Representations}
\textbf{Volumetric maps \& semantic maps.}
In robotics, various map representations have been developed to effectively model the environment.
Volumetric representations, such as voxel-based methods, are widely used for maintaining information like occupancy~\cite{hornung2013octomap} or signed distance fields~\cite{FIESTA, oleynikova2017voxblox}.
Metric-semantic maps combine geometric and semantic information, offering actionable representations of environments.
Common forms include semantics-augmented occupancy maps~\cite{asgharivaskasi2023semantic, dang2018autonomous}, object-based semantic maps~\cite{liu2024slideslam}, and 3D scene graphs~\cite{armeni20193d, wu2021scenegraphfusion, looper20233d, hughes2024ijrr}.
Scene graphs are useful for capturing semantic concepts and relationships through a compact representation~\cite{armeni20193d, wu2021scenegraphfusion, looper20233d}.
Recent work extend hierarchical representations with language features for open-vocabulary scene understanding~\cite{maggio2024clio,devarakonda2024orionnav,conceptgraphs,werby2024hierarchical}.
However, these methods are limited to structured environments, prompting the need for a balance between sparse hierarchical representations and dense feature embeddings for unstructured environments.

\begin{figure*} [!t]
    \centering
    \includegraphics[width=0.95\linewidth]{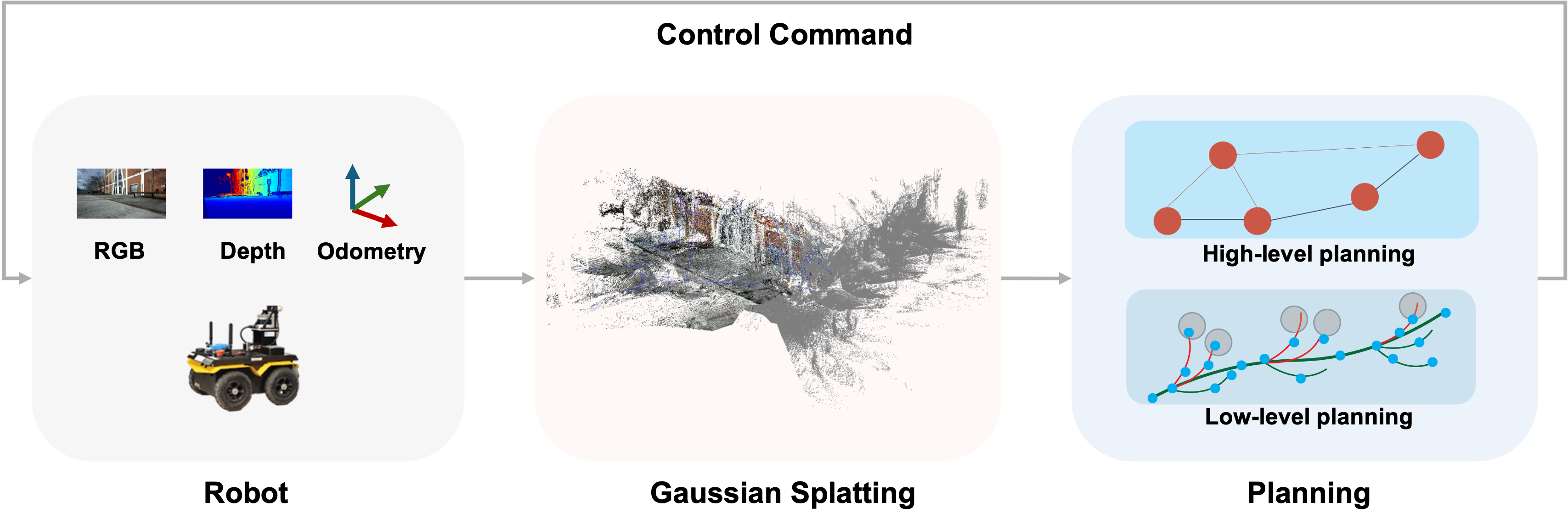}
    \caption{Overview of the autonomy framework.}
    \label{fig:autonomy_overview}
\end{figure*}

\textbf{Gaussian splatting.}
Learned map representations from the computer vision field, including Neural Radiance Fields (NeRF)~\cite{mildenhall2021nerf} and 3D Gaussian Splatting (3DGS)~\cite{kerbl20233dgs}, have become more common in robotics applications.
The concept of 3D Gaussian splatting, as introduced in \cite{kerbl20233dgs}, offers a unique method for capturing geometry using Gaussians while incorporating color and opacity information, enabling high-quality scene rendering.
Building on this, several approaches construct these maps in real time using color and depth measurements within Simultaneous Localization and Mapping (SLAM) frameworks.
The work proposed in \cite{keetha2024splatam} utilizes silhouette masks for efficient rendering optimization.
The work in \cite{peng2024rtg} handles Gaussian parameters differently for color and depth rendering, resulting in a more memory-efficient scene representation, and \cite{hu2025cg} focuses on estimating stability and uncertainty for efficient scene representation. 
The work in \cite{yu2025hammer} presents a framework for constructing semantic Gaussian splatting maps from multiple agents.
Gaussian splatting has also emerged as a promising representation of the environment in the context of active mapping \cite{jiang2024multimodal, jiang2023fisherrfactiveviewselection}, which tries to minimize the uncertainty of the map (and in some cases the state of the robot) by optimizing the sequence of actions taken by the robot.
Typically, the exact measure of the reduction in uncertainty is challenging to estimate exactly; this work advocates for a method of uncertainty quantification that has not been leveraged in previous works. 
Additionally, numerous methods \cite{qin2024langsplat} integrate language-embedded 3DGS for scene understanding, typically extracting 2D language features from models like CLIP \cite{radford2021learning} and distilling them into 3D Gaussians. To address mapping of larger scenes in a SLAM setting, the work in \cite{zhu2024loopsplat, yugay2023gaussian} incrementally create submaps and only optimize Gaussians in the latest submap.
The work presented in this paper explores the capabilities of Gaussian splatting for autonomy in unstructured environments.

\subsection{Task-driven Navigation}
\textbf{Information-driven exploration.}
Autonomous exploration describes the process of a robot that generates plans to actively construct a map of an unknown environment.
While both frontier-based exploration~\cite{yamauchi1997frontier, shen2012indoor, zhou2021fuel, yuezhantao2023seer} and information-driven exploration~\cite{saulnier2020information, LukasIG, charrow2015information, bircher2016receding, he2024active} have been widely studied, the definition of frontiers is tightly coupled with the grid-based representation such as occupancy maps and Signed Distance Field (SDF) maps. With the recent advances in the learned map representations, the concept of information-driven exploration has been adopted to evaluate the information~\cite{jiang2023fisherrfactiveviewselection, he2023active, jin2024gsplanner, xu2024hgs}. However, most of the existing approaches still require the existence of a grid-based representation for navigation goal generation, the evaluation of information, or the subsequent planning. Unlike these existing frameworks, our work proposes a unified framework for mapping and planning for information-driven exploration using Gaussian splitting as the representation.

\textbf{Semantic navigation.}
Semantic navigation and task-driven planning addresses the challenge of completing semantically-driven tasks, such as efficiently locating objects. Recent work \cite{dai2024optimalscenegraphplanning} use hierarchical scene graphs with Large Language Models (LLMs) for semantic search, while others use LLMs to aid in frontier selection \cite{jiang2024multimodal}. Other approaches such as \cite{ravichandran2022hierarchical} utilize reinforcement learning with scene graphs for end-to-end navigation policies. In \cite{yokoyama2024vlfm}, Vision-Language Models (VLMs) are used to evaluate frontiers by task relevancy, while \cite{ravichandran2024spine} employed LLMs to select robot behaviors minimizing task completion time. Unlike these approaches that focus primarily on task completion with minimal mapping, our work emphasizes creating dense, reusable maps applicable across various tasks.

\section{Gaussian splatting for autonomy}

This section summarizes our past work  \cite{ong2025atlasnavigatoractivetaskdriven, tao2024rtguiderealtimegaussiansplatting} utilizing Gaussian splatting as a unified representation for navigation. We highlight key capabilities that leverage the strengths of Gaussian splatting to enable autonomous exploration, mapping and planning in unstructured outdoor environments.

\subsection{Gaussian splatting}
Gaussian splatting represents scenes with 3D Gaussians. Each Gaussian has associated parameters position, orientation, scale, opacity, and color. In this work, we also embed language features as additional Gaussian parameters. The differentiable rendering process projects these 3D Gaussians into 2D space efficiently by leveraging the GPU, enabling real-time radiance field construction and rendering. For a given camera pose, corresponding RGB, depth and language feature images can be rendered from the scene. The Gaussian parameters are updated by computing a loss against the original observations.
Unlike NeRFs and other implicit radiance field representations, 3DGS offers an explicit representation that is well-suited for real-time mapping applications. This explicit nature makes 3DGS particularly advantageous for tasks such as collision avoidance, where immediate and direct access to spatial information is crucial.

\subsection{Robot autonomy overview}
We adopt the traditional mapping and planning pipeline for autonomy. The mapping module integrates visual and pose measurements into the Gaussian splatting map. A task-dependent metric is used to identify areas in the map with high utility. The high-level planning module computes a path based on cost-benefit analysis of utility and distance for the robot to follow. The low-level planning module determines safe trajectories through the Gaussians. An overview of such a system is presented in Fig.~\ref{fig:autonomy_overview}. More details of such systems are presented in \cite{tao2024rtguiderealtimegaussiansplatting,ong2025atlasnavigatoractivetaskdriven}.

\begin{figure*}
    \centering
    \includegraphics[width=0.95\linewidth]{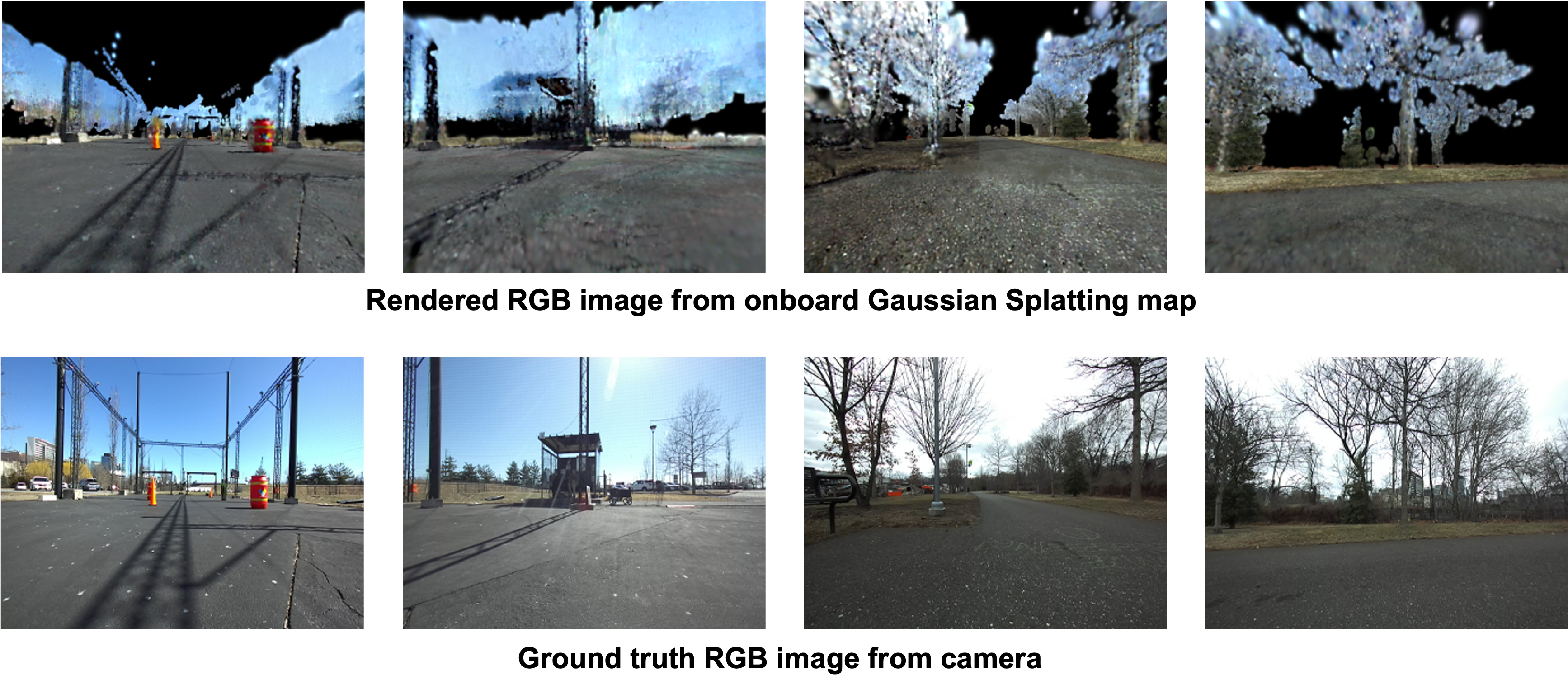}
    \caption{Novel-view rendering from the onboard Gaussian splatting map and the corresponding groundtruth observations.
    }
    \label{fig:rtguide_render}
\end{figure*}

\textbf{Platform. } We develop this autonomy system for the Clearpath Jackal robot.
It is equipped with an AMD Ryzen 5 3600 processor and RTX 4000 Ada SFF graphics card.
The robot's perception system includes an Ouster OS1 LiDAR and a ZED 2i stereo camera.

\subsection{Submapping for large-scale navigation}
As the robot navigates the environment and incrementally builds a map of Gaussians, the number of Gaussians can quickly grow to the order of millions of Gaussians. This poses a problem for both storage of the map on the GPU and querying the Gaussians for collision checking. To this end, we incorporate submapping into our mapping system. Particular considerations are made to enable this submapping structure to work seamlessly with the Gaussian splatting mapping framework. The submapping framework addresses two objectives -- submap-level pose graph updates and memory-efficient mapping.

\textbf{Submap management.}
Submaps contain an anchor pose and the parameters of the associated Gaussians. The positions of the Gaussians are stored relative to the anchor pose of the corresponding submap. A submap is initialized if there are no existing submap anchors within a radius of the robot's current position. When updating the map with SLAM, it is difficult to update every pose since this may cause significant inconsistencies in the shifted Gaussians. Instead, we perform updates on the submap-level. This maintains the geometric consistency of the Gaussians within each submap.

\begin{figure}
\centering
  \begin{tikzpicture}[scale=0.95,trim axis left, trim axis right]
    \pgfmathsetlengthmacro\MajorTickLength{
      \pgfkeysvalueof{/pgfplots/major tick length} * 0.05
    }
    \pgfplotsset{every tick label/.append style={font=\footnotesize}}
    \begin{axis}[ylabel={Time (ms)},
                 xlabel={Number of Gaussians},
                 xmode=log,
                 ymode=log,
                 major tick length=\MajorTickLength,
                 grid=both,
                 label style={font=\footnotesize},
                 grid style={line width=.1pt, draw=gray!10},
                 major grid style={line width=.2pt,draw=gray!50},
                 every tick/.style={
                    black,
                    semithick
      }]

    \addplot table[x index=0,y index=1,col sep=comma] {collision-test-time.dat};
    \addplot+[forget plot,only marks] 
    plot[error bars/.cd, y dir=both, y explicit]
    table[x index=0, y index=1, y error index=2, col sep=comma] {collision-test-time.dat}; 
    \end{axis}
    
    \end{tikzpicture}
\vspace{-0.8cm}
\caption{Time required to test for collisions of trajectories with a 5m horizon (140 points) with respect to the number of Gaussians.
}
\label{fig:collision}
\end{figure}
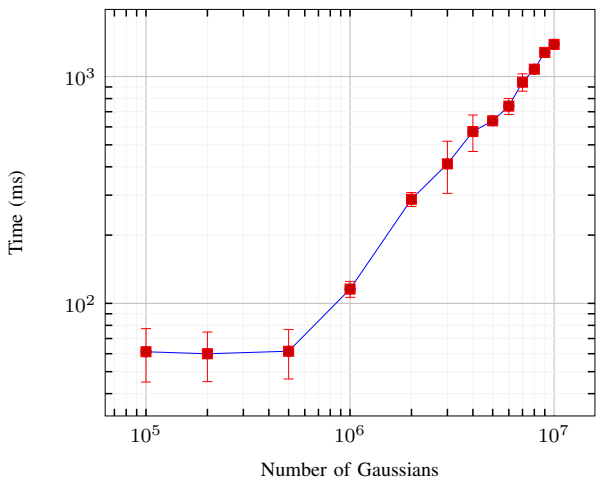

\textbf{Local map management.}
To ensure that the number of Gaussians on GPU remains manageable, the submaps are dynamically loaded and unloaded as the robot moves. 
To ensure safe and efficient navigation, we compute a local map radius which depends on both the perception range of the sensors and also the minimum distance from the robot at which Gaussians must be considered for collision.
Any submaps within the local map radius from the robot are loaded on GPU, while the rest are off-loaded to the CPU. 
This significantly reduces the GPU memory requirement to maintain the map as well as the computation required for collision checking. In Tab.~\ref{table:memory}, we show that our method allows for memory-efficient Gaussian splatting in large environments where other methods fail due to excessive memory requirements.
Methods that do not support submapping are indicated by -- and are not evaluated with submaps.
Similarly, $\times$ indicates that the method failed due to excessive memory requirements.
Gaussian-SLAM and LoopSplat are evaluated with submaps.
Additionally, we show in Fig.~\ref{fig:collision} the importance of managing the number of Gaussians to enable fast collision checking for real-time operation.

\begin{table}[t]
\centering
\caption{\textbf{Comparison of memory performance on scene 00824 of the HM3D dataset. Results from \cite{ong2025atlasnavigatoractivetaskdriven}.}}
\begin{tabular}{
P{0.135\textwidth}|
P{0.08\textwidth}|
P{0.08\textwidth}|
P{0.085\textwidth}}
\hline
\multirow{2}{*}{\textbf{Method}} & \multicolumn{3}{c}{\textbf{{Memory Allocated / Reserved (GB)}}} \\
\cline{2-4}
& 2m submap & 5m submap & No submap \\
\hline
Gaussian-SLAM~\cite{yugay2023gaussian} & \textbf{5.71} / 36.21 & $\times$ & $\times$ \\
LoopSplat~\cite{zhu2024loopsplat} & 9.03 / 21.33 & $\times$ & $\times$ \\
SplaTAM~\cite{keetha2024splatam} & -- & -- & \textbf{13.36} / \textbf{16.32} \\
ATLAS~\cite{ong2025atlasnavigatoractivetaskdriven} & 8.28 / \textbf{9.68} & \textbf{9.77} / \textbf{12.51} & \textbf{13.36} / \textbf{16.32} \\
\hline
\end{tabular}
\bigskip
\label{table:memory}
\vspace{-0.5cm}
\end{table}

\subsection{Collision avoidance in a Gaussian splatting map}
\label{sec: collision_avoidance}
While representations such as voxel grids have been popular options for detecting obstacles due to the discretization of the map~\cite{GroundGrid}, we show that the GPU can be leveraged for fast collision checking in a dense representation~\cite{tao2024rtguiderealtimegaussiansplatting}.
Other work on collision checking in Gaussian Splatting such as~\cite{chen2025splat} rely on the construction of safe flight corridors.
Instead, we find that a exhaustive check in the local map is sufficient.
We use a motion primitive library~\cite{liu2018search} to generate collision-free trajectories through the Gaussian map and develop a GPU-accelerated approach for checking collision against all the Gaussians in the local map.
In particular, in the traditional process of growing the tree of motion primitives, collision checks for each motion primitive are carried out sequentially.
During the process of generating each primitive, test points are sampled, and collision checks are conducted between each test point and obstacles in the map.
To accelerate this process, we first generate all motion primitives at the current level of the search tree and sample all test points from them.
Subsequently, only one collision check is invoked between all test points with all Gaussians that are being considered.
The result of the collision test for each motion primitive can be obtained by executing a logical OR operation for test points that belong to the primitive. 

\section{Information-driven exploration}
Unlike traditional mapping representations, the Gaussian map does not explicitly encode occluded or free space.
Instead of relying on frontiers of unobserved regions, we propose in RT-GuIDE~\cite{tao2024rtguiderealtimegaussiansplatting} an efficient metric for estimating uncertainty in the Gaussian splatting map to determine areas that should be explored next.
\subsection{Uncertainty estimation}
We calculate the change in Gaussian parameters during each mapping optimization and use the magnitude of these parameter updates as an estimate of uncertainty.
This simple metric reveals an interesting behaviour that balances exploration and exploitation.
Gaussians with limited observations or those at the edges of observed regions exhibit higher uncertainty, which supports exploration.
In contrast, areas of the map that have been visited but contain unstable Gaussians necessitate revisitation, which encourages exploitation.

\begin{table}[!t]
\centering
\caption{\textbf{Geometric exploration and mapping results using \cite{tao2024rtguiderealtimegaussiansplatting}.}}
\begin{tabular}{c|c|c|c|c}
\hline
\multirow{2}{*}{\textbf{Methods}}  & \multirow{2}{*}\textbf{PSNR} $\uparrow$ & \multirow{2}{*}\textbf{SSIM} $\uparrow$ & \multirow{2}{*}\textbf{LPIPS} $\downarrow$ & \multirow{2}{*}\textbf{RMSE} $\downarrow$ \\
& [dB] & & & [m] \\
\hline
Ensemble & 11.02 & 0.351 & 0.710 & 1.897 \\
FisherRF~\cite{jiang2023fisherrfactiveviewselection} & 17.18 & 0.655 & 0.408 & 0.676 \\
RT-GuIDE~\cite{tao2024rtguiderealtimegaussiansplatting} & \textbf{18.63} & \textbf{0.736} & \textbf{0.319} & \textbf{0.415} \\
\hline
\end{tabular}
\label{table:rtguide_results}
\end{table}

\begin{figure*}[!ht]
    \centering
    \includegraphics[width=0.98\linewidth,trim={0.5cm 5.0cm 1cm 5.5cm},clip]{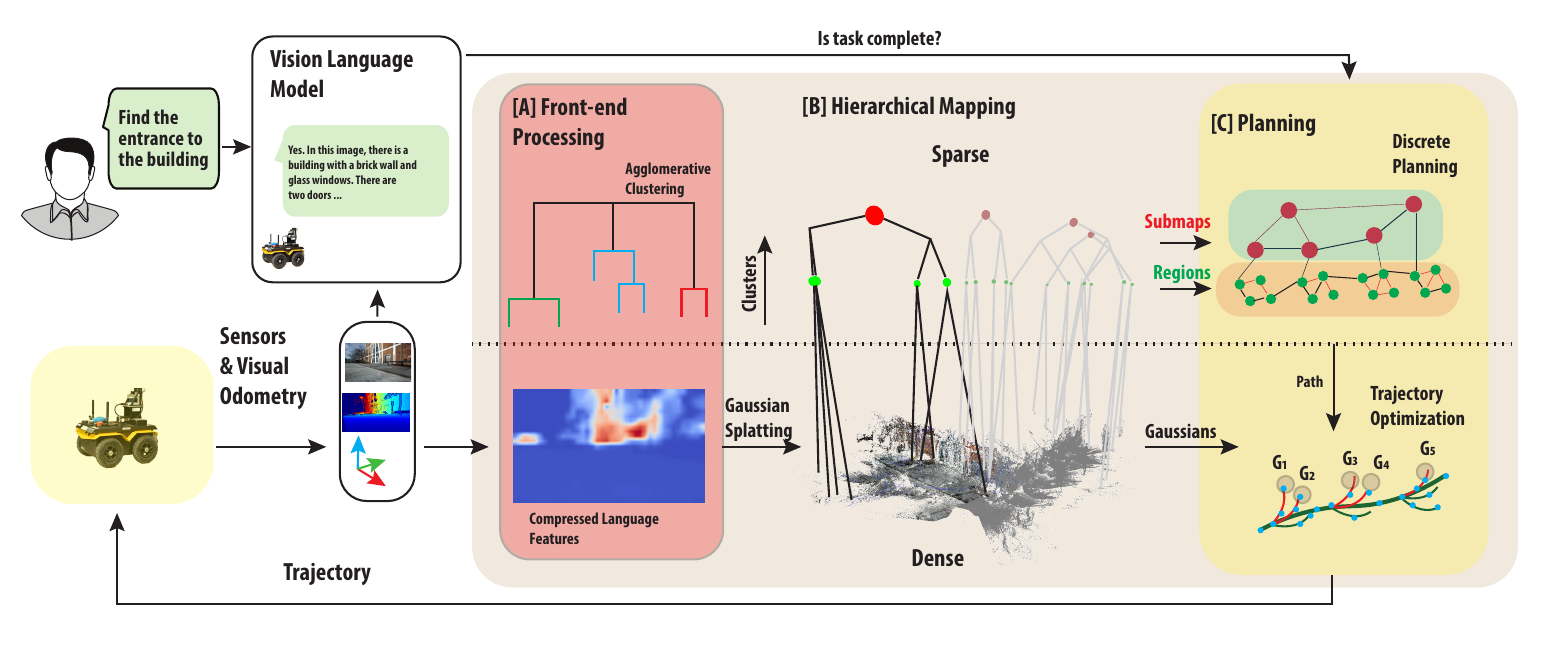}
    \captionof{figure}{An overview of the ATLAS Navigator system. The front-end processing modules [A] extracts and compresses dense pixel-level language features from the image.
    The module also clusters features based on geometry and semantics in the map.
    [B] The hierarchical mapping module fuses RGB and depth images and the odometric path from the robot into the map. 
    The map is partitioned from bottom-up into the objects, regions and submaps.
    The local map comprises the loaded submaps while the rest are unloaded for memory efficiency (shown in gray).
    [C] The planning module consists of a discrete planner that operates on the sparse map and generates a reference path, while the continuous planner computes trajectories through the dense Gaussians in the local map.
    A VLM is queried for task termination.
    Figure from \cite{ong2025atlasnavigatoractivetaskdriven}.
    }
    \label{fig:atlas-system-diagram}
\end{figure*}

\subsection{Planning}
\label{sec:rtguide_planning}
We uniformly sample regions in the map to evaluate their utility, which is defined as the cost-benefit of traveling to the region. 
The region with the highest utility will be set as the navigation goal for the subsequent planning problem.
We solve the planning problem in a hierarchical way by formulating it as a high-level and low-level planning problem.
The goal of the high-level planner is to provide a guidance path that allows us to transit to a high utility region in the space without the need to plan long-range trajectories with computationally expensive collision checks.
We obtain a high-level path using Dijkstra's to find the minimum-cost path.
Subsequently, the low-level planner selects a local navigation goal along this path and generates collision-free and dynamically-feasible trajectory candidates to the local goal with the method presented in Sec.~\ref{sec: collision_avoidance}.
We evaluate the accumulated information along these trajectories and select the candidate with the highest information gain.

\subsection{Experiments \& Results}
We conduct several real-world experiments in both indoor and outdoor environments and show that our proposed information metric facilitates efficient exploration. We compare our approach against two baseline methods - an ensemble-based approach that computes the variance across three models, and FisherRF~\cite{jiang2023fisherrfactiveviewselection} which evaluates viewpoints based on Fisher Information.
We evaluate map reconstruction quality with the following metrics: Peak Signal-to-Noise Ratio (PSNR), Structural Similarity Index \cite{1292216} (SSIM), Learned Perceptual Image Patch Similarity \cite{zhang2018perceptual} (LPIPS) on the color images and Root Mean-Square-Error (RMSE) on the depth images.
Our proposed metric is efficient to compute compared to other methods and lends itself to real-time operation.
The quantitative results are presented in Tab.~\ref{table:rtguide_results}.
In Fig.~\ref{fig:rtguide_render}, we show qualitative results of novel-view rendering from the onboard Gaussian splatting map and the corresponding groundtruth observations of the experiment areas.
We show that our approach is able to construct maps of higher quality and fidelity compared to other information metrics built on frontier-based exploration.

\section{Large-scale Semantic Navigation}
\label{sec:real_world_experiment}

Our work in ATLAS Navigator~\cite{ong2025atlasnavigatoractivetaskdriven} shows that the Gaussian splatting representation can be used to address the challenge of task-oriented navigation. In particular, we demonstrate that we can use Gaussian splatting to incrementally build a rich semantic map and simultaneously identify regions in the map with high relevance for task-driven navigation.
An overview of the system is illustrated in Fig.~\ref{fig:atlas-system-diagram}.

\subsection{Hierarchical Semantic Mapping}
To build and act on semantically rich maps that support diverse tasks in unstructured environments, we require a representation that integrates dense semantic and geometric information which can be constructed and used in real-time for large-scale environments. 
In~\cite{ong2025atlasnavigatoractivetaskdriven}, we propose an agglomerative data structure built on language-embedded Gaussian splatting that does not rely on explicit semantics or geometric priors. 

\textbf{Feature embedding.}
For each observation, we extract CLIP features from the image. To enable efficient embedding of language features in the Gaussian map, we use PCA to compress the features. We obtain representative basis vectors of CLIP features by running PCA on features extracted from the COCO~\cite{lin2014microsoft} 2017 dataset. This is done offline and allows us to just perform a simple projection to compress the features at runtime. These compressed features are then embedded in the map through the differential rendering process. In our experiments, we approximate 512-dimension CLIP features with 24 PCA components.

\textbf{Metric-semantic clustering.}
To efficiently query the language-embedded Gaussians for task relevance and planning, we construct a sparse, hierarchical semantic map from the dense Gaussian map. We perform agglomerative clustering of the Gaussians by Euclidean distance and cosine similarity of the compressed features. This approach allows us to cluster Gaussians without the need for image segmentation as in \cite{werby2024hierarchical,conceptgraphs,jatavallabhula2023conceptfusion}, which may not provide meaningful priors, both geometrically and semantically, in unstructured environments. For a given task, we compute relevancy of the Gaussians clusters and propagate the utility up the tree, providing a sparse semantic representation for planning.

\subsection{Planning}
We build upon the planning approach in Sec.~\ref{sec:rtguide_planning} but instead aim to maximize task relevancy while respecting a given budget. We use the hierarchical structure provided by the submaps and agglomerative clustering to first find a path through the high-level graph of submaps, followed by a more specific path through the object-level graph that visits the submaps in order.

\subsection{Experiments \& Results}
\begin{figure}
    \centering
    \includegraphics[height=5cm,width=0.95\linewidth]{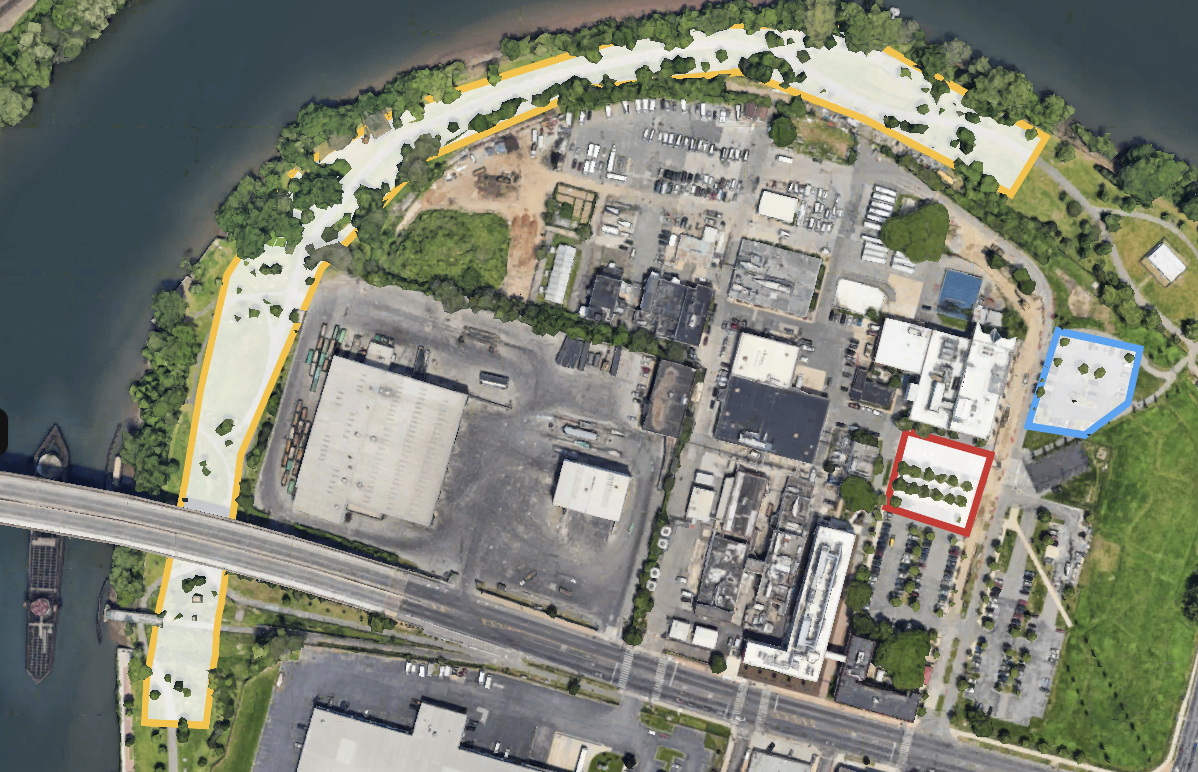}
    \caption{Experiment areas for large-scale semantic navigation.
    The park experiments were conducted in the highlighted yellow areas. The parking lots are highlighted in red and blue.
    Figure from \cite{ong2025atlasnavigatoractivetaskdriven}.}
    \label{fig:experiment-areas}
\end{figure}

We conduct experiments in large-scale, outdoor environments.
The experiment areas are illustrated in Fig.~\ref{fig:experiment-areas}.
In each experiment, we specify the task in natural language.
In \textit{Outdoor1}, we evaluate the framework's capacity to load a pre-built map, localize itself and then navigate to a desired target with the existing graph.
In the rest of the experiments, we showcase the method's applicability to semantic navigation in entirely unexplored environments by leveraging utility signals for task completion.
A VLM determines task termination using a yes/no prompt.

The results are presented in Tab.~\ref{tab:atlas_expts}.
We assess the robot's performance by analyzing its trajectory in each trial.
The actual path length (\textit{PL}) is compared against two benchmark measurements: the shortest possible path (\textit{SP}) and the ground truth path (\textit{GT}).
We calculate \textit{SP} using the completed planning graph, which eliminates odometry discrepancies as both the traversed route and the graph are derived from identical odometry data.
The \textit{GT} path length is determined using human-annotated GPS coordinates on Google Earth.
For each experiment, we compute the competitive ratio (labeled \textit{ratio} in Tab.~\ref{tab:atlas_expts}), comparing the robot's visually-estimated travel distance to these reference measurements.
This ratio is expressed as $\frac{SP}{PL}$ and $\frac{GT}{PL}$.
To provide context for the experiment scale, we report the approximate operational area (derived from Google Earth) and the total number of Gaussians in each map.

\begin{table*}[t]
\begin{center}
\centering
\caption{\textbf{Overview of semantic navigation experiments. \textit{PL} refers to path length, \textit{SP} refers to the shortest path computed on the full planning graph after the experiment is complete, and \textit{GT} refers to the ground truth. Table from \cite{ong2025atlasnavigatoractivetaskdriven}.}}
\label{tab:atlas_expts}
\begin{tabular}{
P{0.08\textwidth}|
P{0.18\textwidth}|
P{0.05\textwidth}|
P{0.06\textwidth}|
P{0.05\textwidth}|
P{0.05\textwidth}|
P{0.05\textwidth}|
P{0.05\textwidth}|
P{0.06\textwidth}|
P{0.07\textwidth}}
\hline
    \multirow{2}{*}{\textbf{Experiment}} & \multirow{2}{*}{\textbf{Task}} & \textbf{Prior} & \multicolumn{5}{c|}{\textbf{Distance (m)}} & {\textbf{Area}} & {\textbf{Num. of}} \\ \cline{4-8}
     &  & \textbf{Map} & {\textbf{PL (m)}} & \multicolumn{2}{c|}{\textbf{SP (m / ratio)}} & \multicolumn{2}{c|}{\textbf{GT (m / ratio)}} & {\textbf{(m$^2$)}} & {\textbf{Gaussians}} \\ \hline 
    Outdoor1 & Navigate to entrance to pier & Yes & 185.78 & -- & -- & 184.86 & 0.99 & 3346.50 & 2,661,562 \\
    Outdoor2 & Inspect road blockage & No & 94.25 & 45.16 & 0.48 & 43.52 & 0.46 & 1063.57 & 1,087,489  \\
    Outdoor3 & Find parking lot & No & 69.19 & 46.439 & 0.67 & 43.22 & 0.62 & 1280.10 & 1,473,420 \\
    Outdoor4 & Find river near road & No & 742.42 & 473.64 & 0.64 & 472.55 & 0.64 & 17791.14 & 7,769,214 \\
    Outdoor5 & Find boardwalk near road & No & 1257.53 & 688.90 &  0.55 & 671.76 &  0.53 & 21870.34 & 11,279,776 \\
    \hline
\end{tabular}
\end{center}
\end{table*}

\section{Lessons Learned}
\label{sec:conclusion}

In this section, we discuss some challenges faced through our experimentation and opportunities to address them in future research.
Specifically, in our past work we augmented the capability of Gaussian splatting to hold semantic features and designed our planner to use the Gaussians as an oracle to check if a point lies in free space.
While this representation is amenable to planning problems, the close coupling of the rendering quality to the captured noisy 3D geometry and pose estimation still proves to be challenging when the planned paths may result in non-smooth behavior on the robot.

\subsection{Gaussian splatting for UAV autonomy}
Implementing a similar framework on a UAV with an embedded computing device such as an NVIDIA Jetson introduces additional challenges.
Work such as~\cite{chen2025splat, jin2025activegs} consider aerial autonomy but note the computational challenges and run the splatting process offboard.
A unified memory architecture may require unloading submaps to disk to maintain large maps.
Additional considerations are required for 3D planning, where computing 3D motion primitives will result in a larger search space and significantly more collision checks.
Alternatively, other planning approaches such as those using safe flight corridors can be adapted to work directly on the Gaussians.
While existing methods for hierarchical Gaussian splatting focus on rendering at different levels of detail, these techniques can potentially be leveraged for efficient collision checking at different resolutions.

\subsection{Dense semantics for scene understanding}
\begin{figure}[!t]
\centering
\begin{subfigure}{0.22\textwidth}
    \centering
    \includegraphics[height=2.5cm]{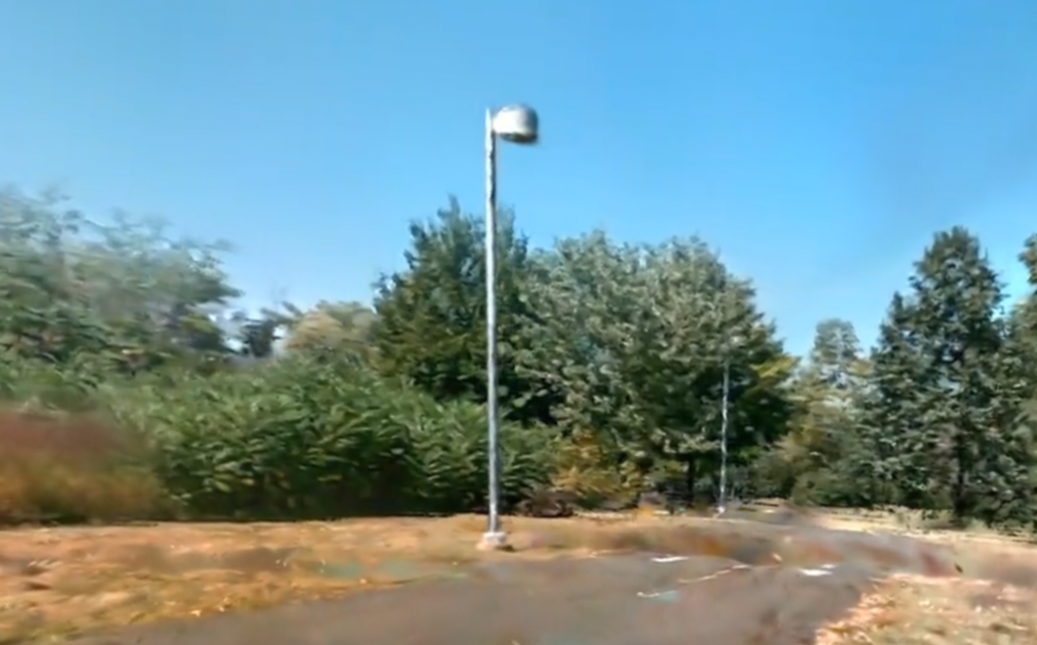}    
\end{subfigure}
\hspace{0.2cm}
\begin{subfigure}{0.22\textwidth}
    \centering
    \includegraphics[height=2.5cm]{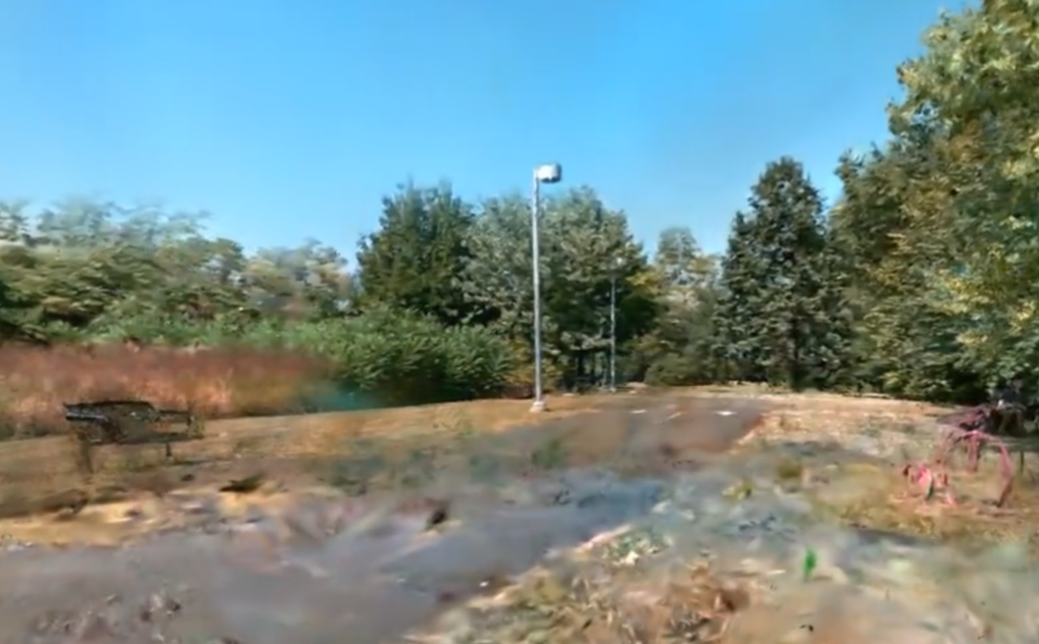}
\end{subfigure}

\caption{Novel-view renderings from a Gaussian splatting map. The geometry of thin obstacles such as this lamppost can be accurately captured with such a representation.}
\label{fig:lamppost}
\end{figure}

While a sparse semantic graph is suited for planning, the dense semantics in the map can be useful for terrain classification and traversability estimation.
Maintaining a dense geometric map in unstructured environments can be advantageous since the geometry of small and thin obstacles such as branches and lampposts can be more accurately represented.
Such an example is shown in Fig.~\ref{fig:lamppost}.
Additionally, having a dense semantic map opens up opportunities for handling multiple tasks and provides more context for executing under-specified tasks.

\subsection{Towards tightly-coupled Gaussian splatting SLAM}
In this work, we adopt a parallel tracking and mapping framework, relying on an external visual-inertial odometry source to construct the map.
In the case of using LiDAR as an odometry source, we noticed significant errors in the map due to misalignment between the LiDAR odometry estimates and the camera images.
Such issues are expected when fusing data from different sensors. In the case of a vision-only system, we also encountered issues with localization due to drift in the visual-inertial odometry. While such an approach is common in voxel-based mapping, a more tightly-coupled localization and mapping method is required for fine-grained mapping, especially when photometric quality is of concern such as in radiance field representations.
While existing Gaussian splatting SLAM approaches have difficulty with large viewpoint changes as is common in real-time operation, they nonetheless provide a valuable starting point for advancing toward real-time SLAM systems suitable for autonomous operation.

\section{Conclusion}
Our work demonstrates that Gaussian splatting is a promising unified representation for autonomous robot navigation in challenging unstructured outdoor environments.
By addressing key challenges in large-scale autonomy, the work presented here opens new possibilities for robots to explore, understand and navigate complex environments.
Further research in the highlighted areas will bring us closer to operating effectively in the diverse and challenging settings of the real world.

\balance
\bibliography{literature}

\end{document}